\documentclass[runningheads]{lib/llncs}
\usepackage{graphicx}
\usepackage{amsmath,amssymb} 
\usepackage{booktabs, multirow}
\usepackage{cite}
\usepackage{hyperref}
\usepackage[strings]{underscore}

\usepackage{url}

\newcommand{\etal}{\emph{et al.}}
\newcommand{\sota}{state-of-the-art }    

\begin{document}

\title{
    GANomaly: Semi-Supervised Anomaly Detection via Adversarial Training
}
\titlerunning{GANomaly}

\author{
  Samet Akcay\textsuperscript{1} \and
  Amir Atapour-Abarghouei\textsuperscript{1} \and
  Toby P. Breckon\textsuperscript{1,2}
}
\authorrunning{S. Akcay \etal}

\institute{
  Department of \{Computer Science\textsuperscript{1}, Engineering\textsuperscript{2}\}, Durham University, UK \\
  \email{
    \{
      \href{mailto:samet.akcay@durham.ac.uk}{samet.akcay},
      \href{mailto:amir.atapour-abarghouei@durham.ac.uk}
           {amir.atapour-abarghouei},
      \href{mailto:toby.breckon@durham.ac.uk}{toby.breckon}
    \}@durham.ac.uk
  }
}

\maketitle

\begin{abstract}
    Anomaly detection is a classical problem in computer vision, namely the determination of the \textit{normal} from the \textit{abnormal} when datasets are highly biased towards one class (normal) due to the insufficient sample size of the other class (abnormal). While this can be addressed as a supervised learning problem, a significantly more challenging problem is that of detecting the unknown/unseen anomaly case that takes us instead into the space of a one-class, semi-supervised learning paradigm. We introduce such a novel anomaly detection model, by using a conditional generative adversarial network that jointly learns the generation of high-dimensional image space and the inference of latent space. Employing encoder-decoder-encoder sub-networks in the generator network enables the model to map the input image to a lower dimension vector, which is then used to reconstruct the generated output image. The use of the additional encoder network maps this generated image to its latent representation. Minimizing the distance between these images and the latent vectors during training aids in learning the data distribution for the normal samples. As a result, a larger distance metric from this learned data distribution at inference time is indicative of an outlier from that distribution --- \textit{an anomaly}. Experimentation over several benchmark datasets, from varying domains, shows the model efficacy and superiority over previous \sota approaches.
    
\end{abstract}

\vspace{-0.1cm}
\section{Introduction} \label{sec:introduction}
\vspace{-0.3cm}
Despite yielding encouraging performance over various computer vision tasks, supervised approaches heavily depend on large, labeled datasets. In many of the real world problems, however, samples from the more unusual classes of interest are of insufficient sizes to be effectively modeled. Instead, the task of anomaly detection is to be able to identify such cases, by training only on samples considered to be \textit{normal} and then identifying these unusual, insufficiently available samples (\textit{abnormal}) that differ from the learned sample distribution of normality. For example a tangible application, that is considered here within our evaluation, is that of X-ray screening for aviation or border security --- where anomalous items posing a security threat are not commonly encountered, exemplary data of such can be difficult to obtain in any quantity, and the nature of any anomaly posing a potential threat may evolve due to a range of external factors. However, within this challenging context, human security operators are still competent and adaptable anomaly detectors against new and emerging anomalous threat signatures.

\begin{figure}[t!]
    \centering
    \includegraphics[width=\linewidth]{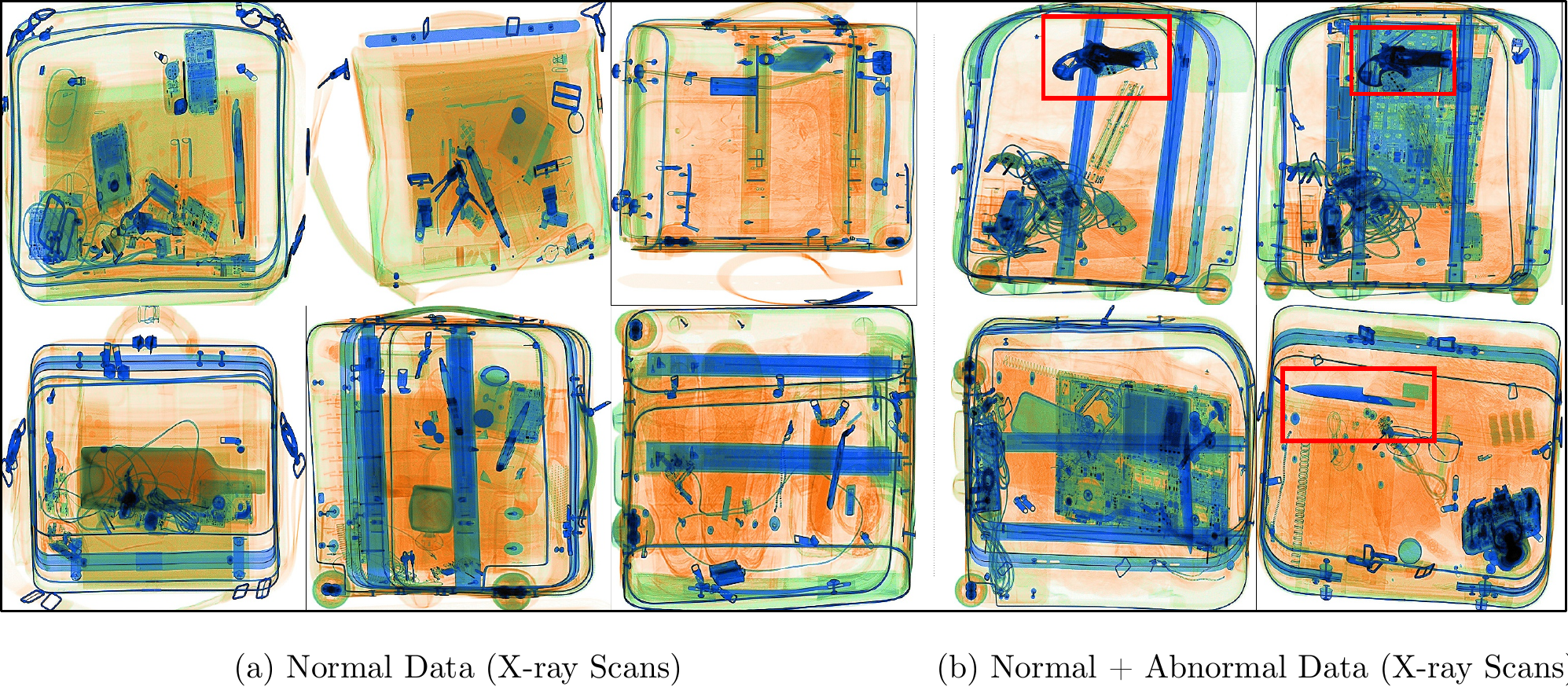}
    \caption{Overview of our anomaly detection approach within the context of an X-ray security screening problem. Our model is trained on normal samples (a), and tested on normal and abnormal samples (b). Anomalies are detected when the output of the model is greater than a certain threshold $\mathcal{A}(x) > \phi$.}
    \label{fig:illustration-of-anomaly-detection}
    \vspace{-0.25cm}
\end{figure}

As illustrated in Figure \ref{fig:illustration-of-anomaly-detection}, a formal problem definition of the anomaly detection task is as follows: given a dataset $\mathcal{D}$ containing a large number of normal samples $\bf X$ for training, and relatively few abnormal examples $\bf \hat X$ for the test, a model $f$ is optimized over its parameters $\bf \theta$. $f$ learns the data distribution $p_{\bf X}$ of the normal samples during training while identifying abnormal samples as outliers during testing by outputting an anomaly score $\mathcal{A}(x)$, where x is a given test example. A Larger $\mathcal{A}(x)$ indicates possible abnormalities within the test image since $f$ learns to minimize the output score during training. $\mathcal{A}(x)$ is general in that it can detect unseen anomalies as being non-conforming to $p_{\bf X}$.

There is a large volume of studies proposing anomaly detection models within various application domains \cite{Abdallah2016,Ahmed2016,Ahmed2016a,Schlegl2017,Kiran2018}. Besides, a considerable amount of work taxonomized the approaches within the literature \cite{Markou2003a,Markou2003,Hodge2004,Chandola2009,Pimentel2014}. In parallel to the recent advances in this field, Generative Adversarial Networks (GAN) have emerged as a leading methodology across both unsupervised and semi-supervised problems. Goodfellow \etal \cite{Goodfellow2014b} first proposed this approach by co-training a pair networks (generator and discriminator). The former network models high dimensional data from a latent vector to resemble the source data, while the latter distinguishes the modeled (i.e., approximated) and original data samples. Several approaches followed this work to improve the training and inference stages \cite{Arjovsky2017,Gulrajani2017}. As reviewed in \cite{Kiran2018}, adversarial training has also been adopted by recent work within anomaly detection.

Schlegl \etal \cite{Schlegl2017} hypothesize that the latent vector of a GAN represents the true distribution of the data and remap to the latent vector by optimizing a pre-trained GAN based on the latent vector. The limitation is the enormous computational complexity of remapping to this latent vector space. In a follow-up study, Zenati \etal \cite{Zenati2018a} train a BiGAN model \cite{Donahue2016}, which maps from image space to latent space jointly, and report statistically and computationally superior results albeit on the simplistic MNIST benchmark dataset \cite{LeCun2010}.

Motivated by \cite{Schlegl2017,Zenati2018a,an2015variational}, here we propose a generic anomaly detection architecture comprising an adversarial training framework. In a similar vein to \cite{Schlegl2017}, we use single color images as the input to our approach drawn only from an example set of $normal$ (non-anomalous) training examples. However, in contrast, our approach does not require two-stage training and is both efficient for model training and later inference (run-time testing). As with \cite{Zenati2018a}, we also learn image and latent vector spaces jointly. Our key novelty comes from the fact that we employ adversarial autoencoder within an encoder-decoder-encoder pipeline, capturing the training data distribution within both image and latent vector space. An adversarial training architecture such as this, practically based on only $normal$ training data examples, produces superior performance over challenging benchmark problems. The main contributions of this paper are as follows:
\begin{itemize}
    \item \emph{semi-supervised anomaly detection} --- a novel adversarial autoencoder within an encoder-decoder-encoder pipeline, capturing the training data distribution within both image and latent vector space, yielding superior results to contemporary GAN-based and traditional autoencoder-based approaches.
    \item \emph{efficacy} --- an efficient and novel approach to anomaly detection that yields both statistically and computationally better performance.
    \item \emph{reproducibility} --- simple and effective algorithm such that the results could be reproduced via the code\footnote{The code is available on \href{https://github.com/samet-akcay/ganomaly}{https://github.com/samet-akcay/ganomaly}} made publicly available.
\end{itemize}

\vspace{-0.1cm}
\section{Related Work} \label{sec:related-work}
\vspace{-0.3cm}
Anomaly detection has long been a question of great interest in a wide range of domains including but not limited to biomedical \cite{Schlegl2017}, financial \cite{Ahmed2016a} and security such as video surveillance \cite{Kiran2018}, network systems \cite{Ahmed2016} and fraud detection \cite{Abdallah2016}. Besides, a considerable amount of work has been published to taxonomize the approaches in the literature \cite{Markou2003a,Markou2003,Hodge2004,Chandola2009, Pimentel2014}. The narrower scope of the review is primarily focused on reconstruction-based anomaly techniques.

The vast majority of the reconstruction-based approaches have been employed to investigate anomalies in video sequences. Sabokrou \etal \cite{Sabokrou2015} investigate the use of Gaussian classifiers on top of autoencoders (global) and nearest neighbor similarity (local) feature descriptors to model non-overlapping video patches. A study by Medel and Savakis \cite{Medel2016} employs convolutional long short-term memory networks for anomaly detection. Trained on normal samples only, the model predicts the future frame of possible standard example, which distinguishes the abnormality during the inference. In another study on the same task, Hasan \etal \cite{Hasan2016} considers a two-stage approach, using local features and fully connected autoencoder first, followed by fully convolutional autoencoder for end-to-end feature extraction and classification. Experiments yield competitive results on anomaly detection benchmarks. To determine the effects of adversarial training in anomaly detection in videos, Dimokranitou \cite{Dimokranitou2017} uses adversarial autoencoders, producing a comparable performance on benchmarks. 


More recent attention in the literature has been focused on the provision of adversarial training. The seminal work of Ravanbakhsh \etal \cite{Ravanbakhsh2017a} utilizes image to image translation \cite{Isola2016} to examine the abnormality detection problem in crowded scenes and achieves \sota on the benchmarks. The approach is to train two conditional GANs. The first generator produces optical flow from frames, while the second generates frames from optical-flow.

The generalisability of the approach mentioned above is problematic since in many cases datasets do not have temporal features. One of the most influential accounts of anomaly detection using adversarial training comes from Schlegl \etal \cite{Schlegl2017}. The authors hypothesize that the latent vector of the GAN represents the distribution of the data. However, mapping to the vector space of the GAN is not straightforward. To achieve this, the authors first train a generator and discriminator using only normal images. In the next stage, they utilize the pre-trained generator and discriminator by freezing the weights and remap to the latent vector by optimizing the GAN based on the $z$ vector. During inference, the model pinpoints an anomaly by outputting a high anomaly score, reporting significant improvement over the previous work. The main limitation of this work is its computational complexity since the model employs a two-stage approach, and remapping the latent vector is extremely expensive. In a follow-up study, Zenati \etal \cite{Zenati2018a} investigate the use of BiGAN \cite{Donahue2016} in an anomaly detection task, examining joint training to map from image space to latent space simultaneously, and vice-versa. Training the model via \cite{Schlegl2017} yields superior results on the MNIST \cite{LeCun2010} dataset.

Overall prior work strongly supports the hypothesis that the use of autoencoders and GAN demonstrate promise in anomaly detection problems \cite{Kiran2018,Schlegl2017,Zenati2018a}. Motivated by the idea of GAN with inference studied in \cite{Schlegl2017} and \cite{Zenati2018a},  we introduce a conditional adversarial network such that generator comprises encoder-decoder-encoder sub-networks, learning representations in both image and latent vector space jointly, and achieving \sota performance both statistically and computationally.
\vspace{-0.1cm}
\section{Our Approach: GANomaly}  \label{sec:method}
\vspace{-0.3cm}

\begin{figure*}[t!]
    \centering
    \includegraphics[width=\textwidth]{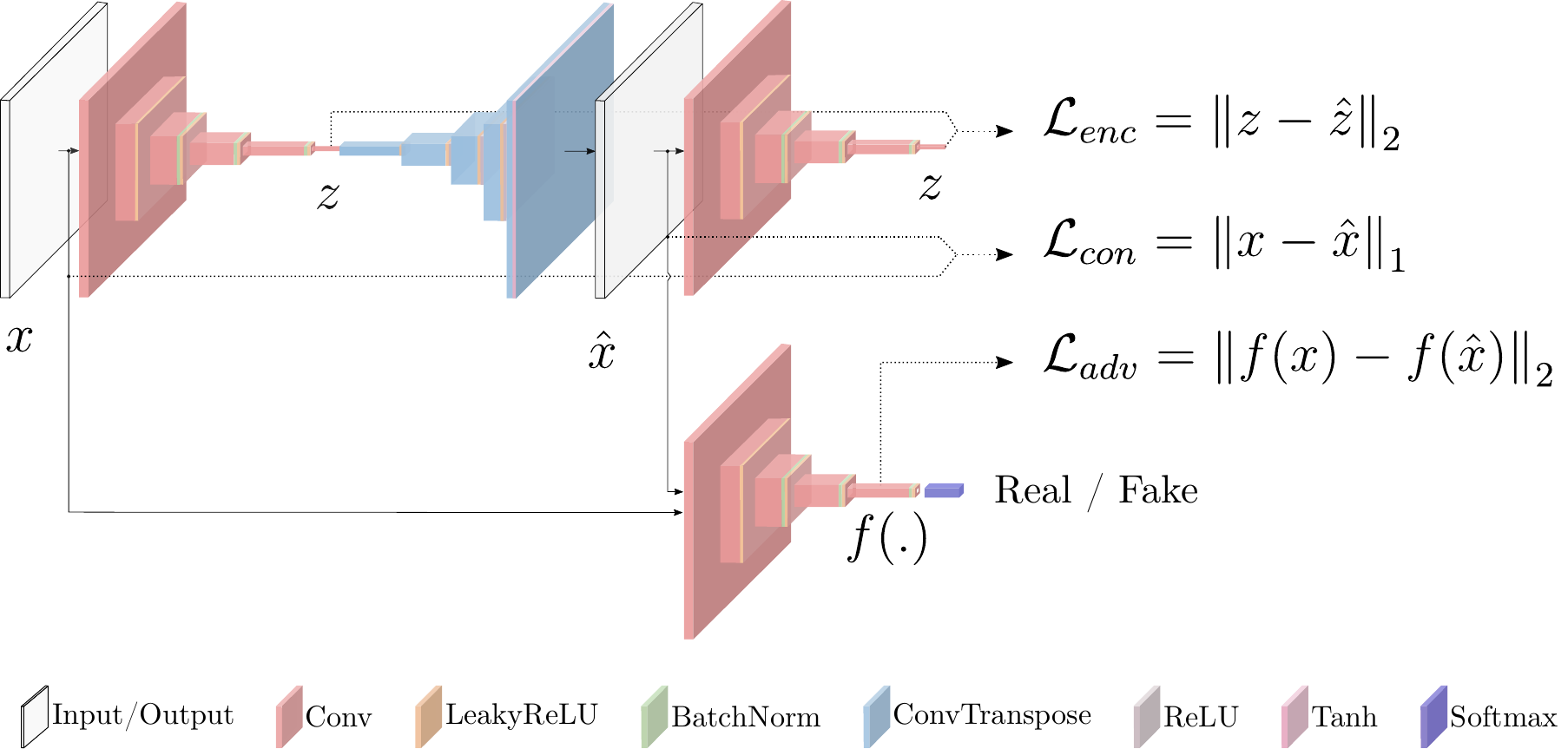}
    \caption{Pipeline of the proposed approach for anomaly detection.}
    \label{fig:pipeline}
\end{figure*} 

To explain our approach in detail, it is essential to briefly introduce the background of GAN.

\subsubsection{Generative Adversarial Networks (GAN)} \label{ssec:gan}
are an unsupervised machine learning algorithm that was initially introduced by Goodfellow et al. \cite{Goodfellow2014b}. The original primary goal of the work is to generate realistic images. The idea being that two networks (generator and discriminator) compete with each other during training such that the former tries to generate an image, while the latter decides whether the generated image is a real or a fake. The generator is a decoder-alike network that learns the distribution of input data from a latent space. The primary objective here is to model high dimensional data that captures the original real data distribution. The discriminator network usually has a classical classification architecture, reading an input image, and determining its validity (i.e., \textit{real vs. fake}).

GAN have been intensively investigated recently due to their future potential \cite{Creswell2017}. To address training instability issues, several empirical methodologies have been proposed \cite{Salimans2016,Arjovsky2017a}. One well-known study that receives attention in the literature is Deep Convolutional GAN (DCGAN) by Radford and Chintala \cite{Radford2015}, who introduce a fully convolutional generative network by removing fully connected layers and using convolutional layers and batch-normalization \cite{Ioffe2015} throughout the network. The training performance of GAN is improved further via the use of Wasserstein loss \cite{Arjovsky2017,Gulrajani2017}.

\subsubsection{Adversarial Auto-Encoders (AAE)} consist of two sub-networks, namely an encoder and a decoder. This structure maps the input to latent space and remaps back to input data space, known as reconstruction. Training autoencoders with adversarial setting enable not only better reconstruction but also control over latent space. \cite{Mirza2014,Makhzani2015,Creswell2017}.

\subsubsection{GAN with Inference} are also used within discrimination tasks by exploiting latent space variables \cite{Chen2016}. For instance, the research by \cite{Creswell2016} suggests that networks are capable of generating a similar latent representation for related high-dimensional image data. Lipton and Tripathi  \cite{Lipton2017} also investigate the idea of inverse mapping by introducing a gradient-based approach, mapping images back to the latent space. This has also been explored in \cite{Dumoulin2016} with a specific focus on joint training of generator and inference networks. The former network maps from latent space to high-dimensional image space, while the latter maps from image to latent space. Another study by Donahue \etal \cite{Donahue2016} suggests that with the additional use of an encoder network mapping from image space to latent space, a vanilla GAN network is capable of learning inverse mapping.

\subsection{Proposed Approach} \label{ssec:proposed-approach}
\subsubsection{Problem Definition.}
Our objective is to train an unsupervised network that detects anomalies using a dataset that is highly biased towards a particular class - i.e., comprising \textit{normal} non-anomalous occurrences only for training. The formal definition of this problem is as follows:

We are given a large tranining dataset $\mathcal{D}$ comprising only $M$ normal images, $\mathcal{D} = \{X_1,\ldots, X_M\}$, and a smaller testing dataset $\hat{\mathcal{D}}$ of N normal and abnormal images, $\hat{\mathcal{D}}=\{(\hat{X}_1, y_1),\ldots, (\hat{X}_N, y_N)\}$, where $y_i\in [0, 1]$ denotes the image label. In the practical setting, the training set is significantly larger than the test set such that $M \gg N$.

Given the dataset, our goal is first to model ${\mathcal{ D}}$ to learn its manifold, then detect the abnormal samples in $\hat{\mathcal{D}}$ as outliers during the inference stage. The model $f$ learns both the normal data distribution and minimizes the output anomaly score $\mathcal{A}(x)$. For a given test image $\hat{x}$, a high anomaly score of $\mathcal{A}(\hat{x})$) indicates possible anomalies within the image. The evaluation criteria for this is to threshold ($\phi$) the score, where $\mathcal{A}(\hat{x}) > \phi$ indicates anomaly.

\subsubsection{Ganomaly Pipeline.} \label{ssec:ganomaly-pipeline}
Figure \ref{fig:pipeline} illustrates the overview of our approach, which contains two encoders, a decoder, and discriminator networks, employed within three sub-networks.


First sub-network is a bow tie autoencoder network behaving as the generator part of the model. The generator learns the input data representation and reconstructs the input image via the use of an encoder and a decoder network, respectively. The formal principle of the sub-network is the following: The generator $G$ first reads an input image $x$, where $x \in \mathbb{R}^{w \times h \times c}$, and forward-passes it to its encoder network $G_E$. With the use of convolutional layers followed by batch-norm and leaky $ReLU()$ activation, respectively, $G_E$ downscales $x$ by compressing it to a vector $z$, where $z \in \mathbb{R}^d$. $z$ is also known as the bottleneck features of $G$ and hypothesized to have the smallest dimension containing the best representation of $x$. The decoder part $G_D$ of the generator network $G$ adopts the architecture of a DCGAN generator \cite{Radford2015}, using convolutional transpose layers, $ReLU()$ activation and batch-norm together with a tanh layer at the end. This approach upscales the vector $z$ to reconstruct the image $x$ as $\hat{x}$. Based on these, the generator network $G$ generates image $\hat{x}$ via $\hat{x}=G_D(z)$, where $z=G_E(x)$.

The second sub-network is the encoder network $E$ that compresses the image $\hat{x}$ that is reconstructed by the network $G$. With different parametrization, it has the same architectural details as $G_E$.
$E$ downscales $\hat{x}$ to find its feature representation $\hat{z}=E(\hat{x})$. The dimension of the vector $\hat z$ is the same as that of $z$ for consistent comparison. This sub-network is one of the unique parts of the proposed approach. Unlike the prior autoencoder-based approaches, in which the minimization of the latent vectors is achieved via the bottleneck features, this sub-network $E$ explicitly learns to minimize the distance with its parametrization. During the test time, moreover, the anomaly detection is performed with this minimization. 

The third sub-network is the discriminator network $D$ whose objective is to classify the input $x$ and the output $\hat x$ as real or fake, respectively. This sub-network is the standard discriminator network introduced in DCGAN \cite{Radford2015}.

Having defined our overall multi-network architecture, as depicted in Figure \ref{fig:pipeline}, we now move on to discuss how we formulate our objective for learning.

\subsection{Model Training} \label{ssec:model-training}
We hypothesize that when an abnormal image is forward-passed into the network $G$, $G_D$ is not able to reconstruct the abnormalities even though $G_E$ manages to map the input $X$ to the latent vector $z$. This is because the network is modeled only on normal samples during training and its parametrization is not suitable for generating abnormal samples. An output $\hat{X}$ that has missed abnormalities can lead to the encoder network $E$ mapping $\hat{X}$ to a vector $\hat{z}$ that has also missed abnormal feature representation, causing dissimilarity between $z$ and $\hat{z}$. When there is such dissimilarity within latent vector space for an input image $X$, the model classifies $X$ as an anomalous image. To validate this hypothesis, we formulate our objective function by combining three loss functions, each of which optimizes individual sub-networks.

\subsubsection{Adversarial Loss.} Following the current trend within the new anomaly detection approaches \cite{Schlegl2017,Zenati2018a}, we also use feature matching loss for adversarial learning. Proposed by Salimans \etal \cite{Salimans2016}, feature matching is shown to reduce the instability of GAN training. Unlike the vanilla GAN where $G$ is updated based on the output of $D$ (\textit{real/fake}), here we update $G$ based on the internal representation of $D$.
Formally, let $f$ be a function that outputs an intermediate layer of the discriminator $D$ for a given input $x$ drawn from the input data distribution $p_{\bf X}$, feature matching computes the $\mathcal{L}_2$ distance between the feature representation of the original and the generated images, respectively. Hence, our adversarial loss $\mathcal{L}_{adv}$ is defined as:

\begin{equation} \label{eq:adversarial-loss}
    \mathcal{L}_{adv}  =  \mathbb{E}_{x \sim p_{\bf X}} {\lVert f(x) - \mathbb{E}_{x \sim p_{\bf X}} f(G(x)\rVert}_2.
\end{equation}

\subsubsection{Contextual Loss.}
The adversarial loss $\mathcal{L}_{adv}$ is adequate to fool the discriminator $D$ with generated samples. However, with only an adversarial loss, the generator is not optimized towards learning contextual information about the input data. It has been shown that penalizing the generator by measuring the distance between the input and the generated images remedies this issue \cite{Isola2016}. Isola \etal \cite{Isola2016} show that the use of $\mathcal{L}_1$ yields less blurry results than $\mathcal{L}_2$. Hence, we also penalize $G$ by measuring the $\mathcal{L}_1$ distance between the original $x$ and the generated images ($\hat{x} = G(x)$) using a contextual loss $\mathcal{L}_{con}$ defined as:

\begin{equation} \label{eq:contextual-loss}
    \mathcal{L}_{con} = \mathbb{E}_{x \sim p_{\bf X}} {\lVert x - G(x) \rVert}_1.
\end{equation}

\subsubsection{Encoder Loss.}
The two losses introduced above can enforce the generator to produce images that are not only realistic but also contextually sound. Moreover, we employ an additional encoder loss $\mathcal{L}_{enc}$ to minimize the distance between the bottleneck features of the input ($z=G_E(x)$) and the encoded features of the generated image ($\hat{z} = E(G(x))$). $\mathcal{L}_{enc}$ is formally defined as

\begin{equation} \label{eq:encoder-loss}
    \mathcal{L}_{enc} = \mathbb{E}_{x \sim p_{\bf X}} {\lVert G_E(x) - E(G(x)) \rVert}_2.
\end{equation}
In so doing, the generator learns how to encode features of the generated image for normal samples. For anomalous inputs, however, it will fail to minimize the distance between the input and the generated images in the feature space since both $G$ and $E$ networks are optimized towards normal samples only.

Overall, our objective function for the generator becomes the following:
\begin{equation} \label{eq:generator-loss}
    \mathcal{L} = w_{adv} \mathcal{L}_{adv} +
                  w_{con} \mathcal{L}_{con} +
                  w_{enc} \mathcal{L}_{enc}
\end{equation}
where $w_{adv}$, $w_{adv}$ and $w_{adv}$ are the weighting parameters adjusting the impact of individual losses to the overall objective function.

\begin{figure}
    \centering
    \includegraphics[width=.95\textwidth]{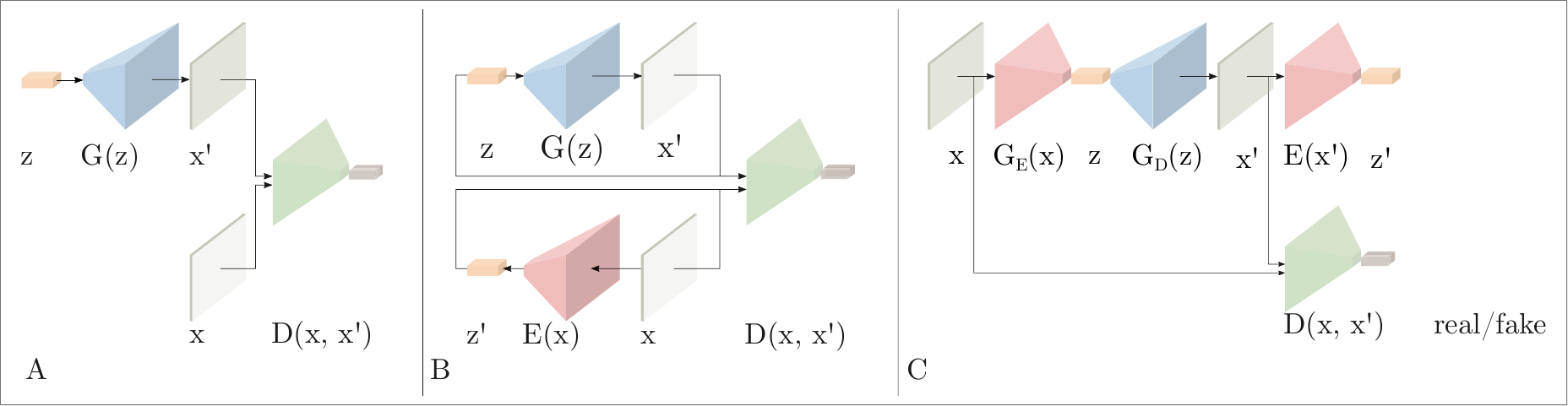}
    \caption{Comparison of the three models. A) AnoGAN \cite{Schlegl2017}, B) Efficient-GAN-Anomaly \cite{Zenati2018a}, C) Our Approach: GANomaly}
    \label{fig:pipeline-comparison}
    \vspace{-0.25cm}
\end{figure}

\vspace{-0.1cm}
\subsection{Model Testing}
During the test stage, the model uses $\mathcal{L}_{enc}$ given in Eq \ref{eq:encoder-loss} for scoring the abnormality of a given image. Hence, for a  test sample $\hat x$, our anomaly score $\mathcal{A}(\hat{x})$ or $s_{\hat{x}}$ is defined as

\begin{equation} \label{eq:anomaly-score}
    \mathcal{A}(\hat{x}) = {\lVert G_E(\hat{x}) - E(G(\hat{x})) \rVert}_1.
\end{equation}

To evaluate the overall anomaly performance, we compute the anomaly score for individual test sample $\hat x$ within the test set $\mathcal{\hat{D}}$, which in turn yields us a set of anomaly scores $\mathcal{S}=\{s_i: \mathcal{A}(\hat{x_i}), \hat{x_i} \in \mathcal{\hat{D}}\}$. We then apply feature scaling to have the anomaly scores within the probabilistic range of $[0, 1]$.

\begin{equation} \label{eq:anomaly-score-scaled}
    s_i'  = \frac{s_i - min(\mathcal{S})}{max(\mathcal{S}) - min(\mathcal{S})}
\end{equation}

The use of Eq \ref{eq:anomaly-score-scaled} ultimately yields an anomaly score vector $\mathcal{S}'$ for the final evaluation of the test set $\mathcal{\hat{D}}$.
\vspace{-0.1cm}
\section{Experimental Setup} \label{sec:experimental-setup} 
\vspace{-0.3cm}


To evaluate our anomaly detection framework, we use three types of dataset ranging from the simplistic benchmark of MNIST\cite{LeCun2010}, the reference benchmark of CIFAR\cite{Krizhevsky2009} and the operational context of anomaly detection within X-ray security screening\cite{Akcay2018}.

\subsubsection{MNIST.}
To replicate the results presented in \cite{Zenati2018a}, we first experiment on MNIST data \cite{LeCun2010} by treating one class being an anomaly, while the rest of the classes are considered as the normal class. In total, we have ten sets of data, each of which consider individual digits as the anomaly.

\subsubsection{CIFAR10.}
Within our use of the CIFAR dataset, we again treat one class as abnormal and the rest as normal. We then detect the outlier anomalies as instances drawn from the former class by training the model on the latter labels.

\subsubsection{University Baggage Anomaly Dataset --- (UBA).}
This sliding window patched-based dataset comprises 230,275 image patches. Normal samples are extracted via an overlapping sliding window from a full X-ray image, constructed using single conventional X-ray imagery with associated false color materials mapping from dual-energy \cite{rogers2016automated}. Abnormal classes ($122,803$) are of 3 sub-classes --- knife ($63,496$), gun ($45,855$) and gun component ($13,452$) --- contain manually cropped threat objects together with sliding window patches whose intersection over union with the ground truth is greater than $0.3$. 

\subsubsection{Full Firearm vs. Operational Benign --- (FFOB).}
In addition to these datasets, we also use the UK government evaluation dataset \cite{CAST2016}, comprising both expertly concealed firearm (threat) items and operational benign (non-threat) imagery from commercial X-ray security screening operations (baggage/parcels). Denoted as FFOB, this dataset comprises $4,680$ firearm full-weapons as full abnormal and $67,672$ operational benign as full normal images, respectively.

The procedure for train and test set split for the above datasets is as follows: we split the normal samples such that $80\%$ and $20\%$ of the samples are considered as part of the train and test sets, respectively. 
We then resize MNIST to $32\times32$, DBA and FFOB to $64\times64$, respectively.

Following Schlegl \etal \cite{Schlegl2017} (AnoGAN)  and Zenati \etal  \cite{Zenati2018a} (EGBAD), our adversarial training is also based on the standard DCGAN approach \cite{Radford2015} for a consistent comparison. As such, we aim to show the superiority of our multi-network architecture regardless of using any tricks to improve the GAN training. In addition, we also compare our method against the traditional variational autoencoder architecture  \cite{an2015variational} (VAE) to show the advantage of our multi-network architecture. We implement our approach in PyTorch \cite{Paszke2017} (v0.4.0 with Python 3.6.5) by optimizing the networks using Adam \cite{Kingma2014} with an initial learning rate $lr=2e^{-3}$, and momentums $\beta_1=0.5$, $ \beta_2=0.999$. Our model is optimized based on the weighted loss $\mathcal{L}$ (defined in Equation \ref{eq:generator-loss})  using the weight values $w_{bce}=1$, $w_{rec}=50$ and $w_{enc}=1$, which were empirically chosen to yield optimum
results. (Figure \ref{fig:hyper-parameter} (b)). We train the model for 15, 25, 25 epochs for MNIST, UBA and FFOB datasets, respectively. Experimentation is performed using a dual-core Intel Xeon E5-2630 v4 processor and NVIDIA GTX Titan X GPU.

\section{Results} \label{sec:results}
We report results based on the area under the curve (AUC) of the Receiver Operating Characteristic (ROC), true positive rate (TPR) as a function of false positive rate (FPR) for different points, each of which is a TPR-FPR value for different thresholds.



\begin{figure}[t!]
    \centering
    \includegraphics[width=\linewidth]{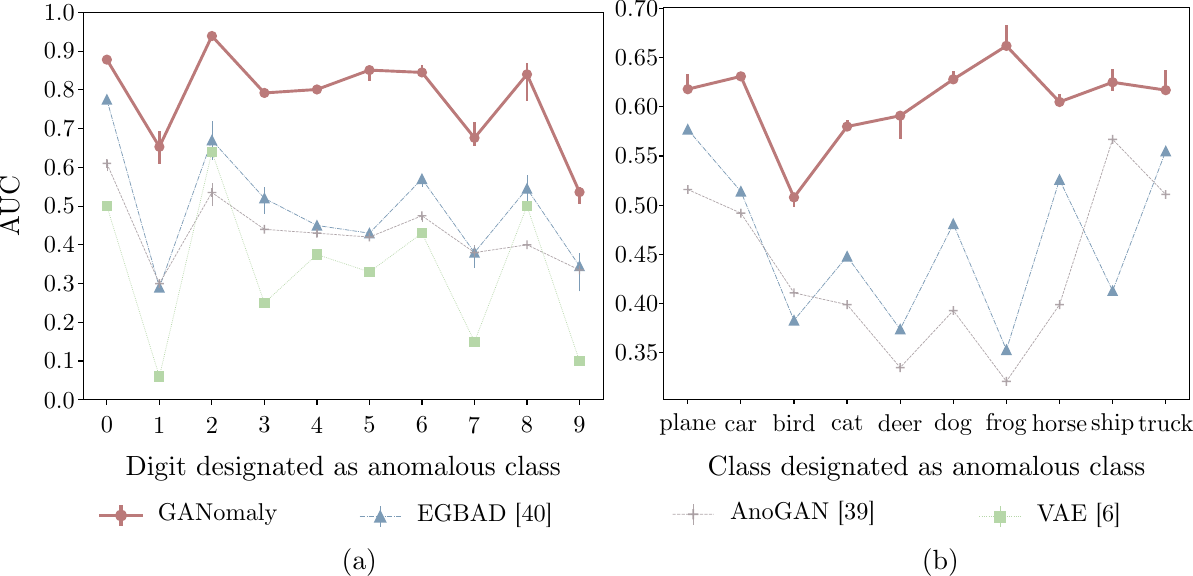}
    \caption{Results for MNIST (a) and CIFAR (b) datasets. Variations due to the use of 3 different random seeds are depicted via error bars. All but GANomaly results in (a) were obtained from \cite{Zenati2018a}. }
    \label{fig:mnist-cifar-results}
\end{figure}

Figure \ref{fig:mnist-cifar-results} (a) presents the results obtained on MNIST data using 3 different random seeds, where we observe the clear superiority of our approach over previous contemporary models \cite{an2015variational,Schlegl2017,Zenati2018a}. For each digit chosen as anomalous, our model achieves higher AUC than EGBAD \cite{Zenati2018a}, AnoGAN\cite{Schlegl2017} and variational autoencoder pipeline VAE \cite{an2015variational}. Due to showing its poor performance within relatively unchallenging dataset, we do not include VAE in the rest of experiments.
Figure \ref{fig:mnist-cifar-results} (b) shows the performance of the models trained on the CIFAR10 dataset. We see that our model achieves the best AUC performance for any of the class chosen as anomalous. The reason for getting relatively lower quantitative results within this dataset is that for a selected abnormal category, there exists a normal class that is similar to the abnormal (plane vs. bird, cat vs. dog, horse vs. deer and car vs. truck).


\begin{table}
    \centering
    \begin{tabular}{@{}lcccccc@{}}  \toprule
                & \multicolumn{4}{c}{UBA}                                           &  & \multicolumn{1}{c}{FFOB} \\ \cmidrule(lr){2-5} \cmidrule(l){7-7}
    Method   & gun            & gun-parts      & knife          & overall        &  & full-weapon              \\ \midrule
    AnoGAN \cite{Schlegl2017}  & 0.598          & 0.511          & \textbf{0.599} & 0.569          &  & 0.703                    \\
    EGBAD \cite{Zenati2018a}  & 0.614          & 0.591          & 0.587          & 0.597          &  & 0.712                    \\
    GANomaly & \textbf{0.747} & \textbf{0.662} & 0.520          & \textbf{0.643} &  & \textbf{0.882}    \\ \bottomrule
    \end{tabular}
    \caption{AUC results for UBA and FFOB datasets}
    \label{tab:xray-results}
\end{table}

For UBA and FFOB datasets shown in  Table \ref{tab:xray-results}, our model again outperforms other approaches excluding the case of the \emph{knife}. In fact,  the performance of the models for \emph{knife} is comparable. Relatively lower performance of this class is its shape simplicity, causing an overfit and hence high false positives. For the overall performance, however, our approach surpasses the other models, yielding AUC of $0.666$ and $0.882$ on the UBA and FFOB datasets, respectively.

\begin{figure}[t!]
    \centering
    \includegraphics[width=\linewidth]{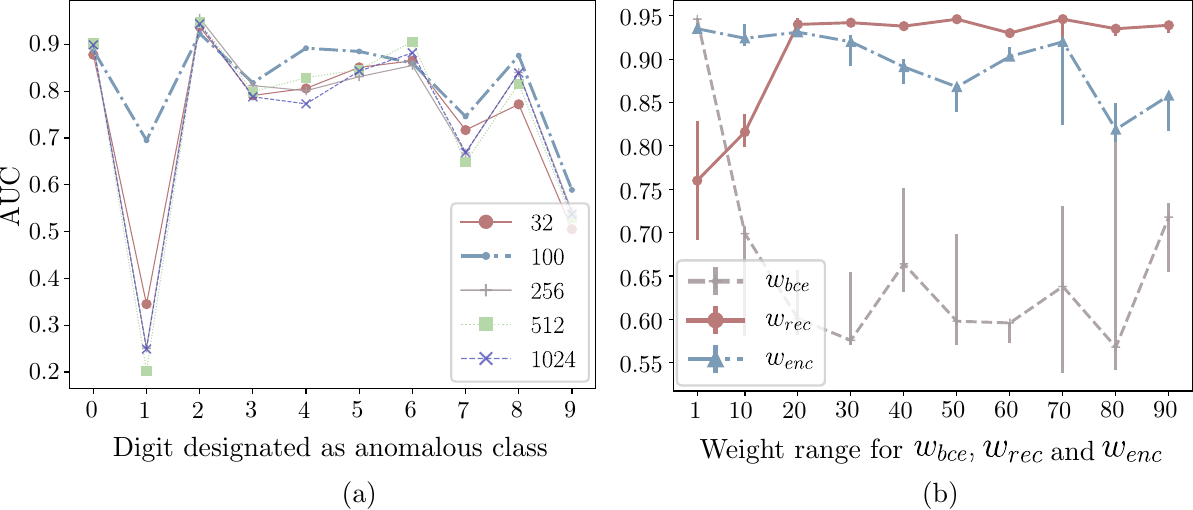}
    \caption{(a) Overall performance of the model based on varying size of the latent vector $z$. (b) Impact of weighting the losses on the overall performance. Model is trained on MNIST dataset with an abnormal digit-2}
    \label{fig:hyper-parameter}
\end{figure}

Figure \ref{fig:hyper-parameter} depicts how the choice of hyper-parameters ultimately affect the overall performance of the model. In Figure \ref{fig:hyper-parameter} (a), we see that the optimal performance is achieved when the size of the latent vector $z$ is $100$ for the MNIST dataset with an abnormal digit-2. Figure \ref{fig:hyper-parameter} (b) demonstrates the impact of tuning the loss function in Equation \ref{eq:generator-loss} on the overall performance. The model achieves the highest AUC when $w_{bce}=1$, $w_{rec}=50$ and $w_{enc}=1$. We empirically observe the same tuning-pattern for the rest of datasets.

\begin{figure}[t!]
    \centering
    \includegraphics[width=\linewidth]{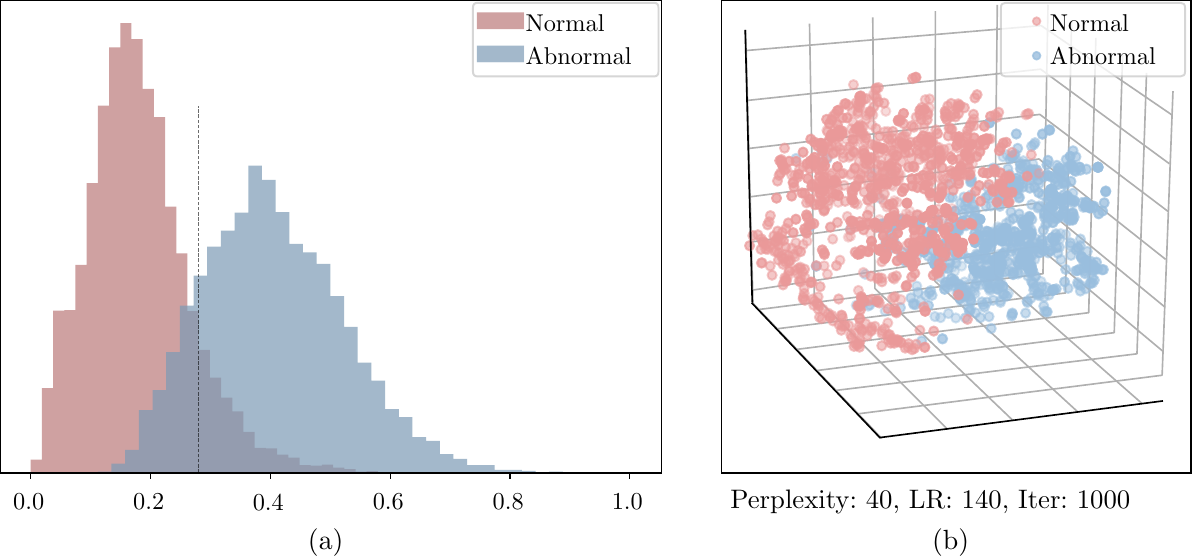}
    \caption{(a) Histogram of the scores for both normal and abnormal test samples. (b) t-SNE visualization of the features extracted from the last conv. layer $f(.)$ of the discriminator}
    \label{fig:histogram-tsne}
\end{figure}

Figure \ref{fig:histogram-tsne} provides the histogram of the anomaly scores during the inference stage (a) and t-SNE visualization of the features extracted from the last convolutional layer of the discriminator network (b). Both of the figures demonstrate a clear separation within the latent vector $z$ and feature $f(.)$ spaces.

Table \ref{tab:runtime} illustrates the runtime performance of the GAN-based models. Compared to the rest of the approaches, AnoGAN \cite{Schlegl2017} is computationally rather expensive since optimization of the latent vector is needed for each example. For EGBAD \cite{Zenati2018a}, we report similar runtime performance to that of the original paper. Our approach, on the other hand, achieves the highest runtime performance. Runtime performance of both UBA and FFOB datasets are comparable to MNIST even though their image and network size are double than that of MNIST.
\begin{table}
    \centering
    \begin{tabular}{@{}lcccc@{}}
    \toprule
    Model    & MNIST & CIFAR & DBA   & FFOB  \\ \midrule
    AnoGAN \cite{Schlegl2017}  & 7120 & 7120 & 7110  & 7223  \\
    EGBAD \cite{Zenati2018a}   & 8.92 & 8.71 & 8.88 & 8.87 \\
    GANomaly & \textbf{2.79} & \textbf{2.21} & \textbf{2.66} & \textbf{2.53} \\ \bottomrule
    \end{tabular}
    \caption{Computational performance of the approaches. (Runtime in terms of millisecond)}
    \label{tab:runtime}
    \vspace{-0.3cm}
\end{table}

\begin{figure}
    \centering
    \includegraphics[width=\linewidth]{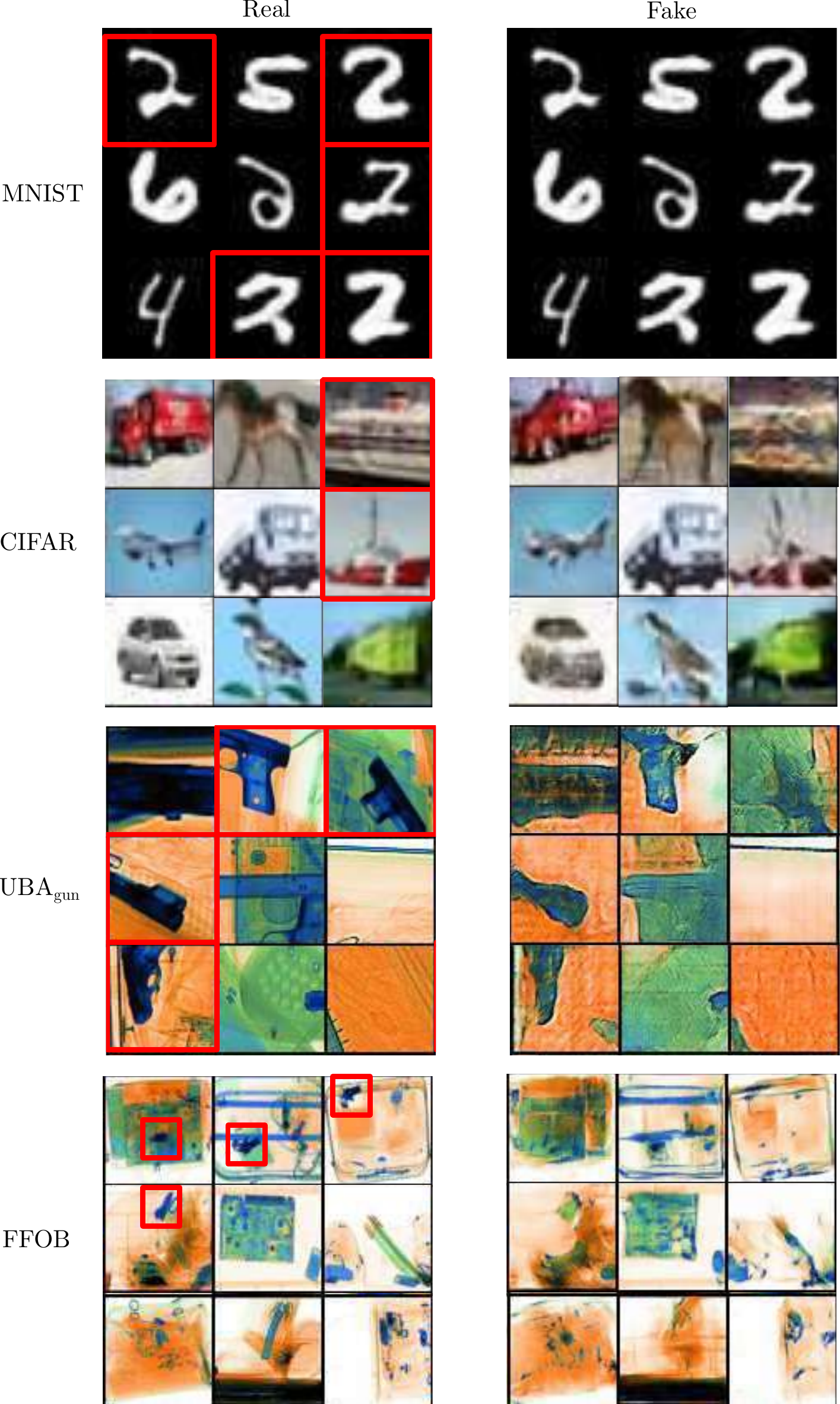}
    \caption{Exemplar real and generated samples containing normal and abnormal objects in each dataset. The model fails to generate abnormal samples not being trained on.}
    \label{fig:output-examples}
\end{figure}

A set of examples in Figure \ref{fig:output-examples} depict real and fake images that are respectively the input and output of our model. We expect the model to fail when generating anomalous samples. As can be seen in Figure \ref{fig:output-examples}(a), this is not the case for the class of 2 in the MNIST data. This stems from the fact that MNIST dataset is relatively unchallenging, and the model learns sufficient information to be able to generate samples not seen during training. Another conclusion that could be drawn is that distance in the latent vector space provides adequate details for detecting anomalies even though the model cannot distinguish abnormalities in the image space. On the contrary to the MNIST experiments, this is not the case. Figures \ref{fig:output-examples} (b-c) illustrate that model is unable to produce abnormal objects.

Overall these results purport that our approach yields both statistically and computationally superior results than leading \sota approaches \cite{Schlegl2017,Zenati2018a}.
\vspace{-0.2cm}
\section{Conclusion} \label{sec:conclusion}
\vspace{-0.3cm}
We introduce a novel encoder-decoder-encoder architectural model for general anomaly detection enabled by an adversarial training framework. Experimentation across dataset benchmarks of varying complexity, and within the operational anomaly detection context of X-ray security screening, shows that the proposed method outperforms both contemporary \sota GAN-based and traditional autoencoder-based anomaly detection approaches with generalization ability to any anomaly detection task. Future work will consider employing emerging contemporary GAN optimizations \cite{Salimans2016,Gulrajani2017,Arjovsky2017a}, known to improve generalized adversarial training.

\bibliographystyle{ref/splncs04}
\bibliography{ref/library.bib}

\begin{thebibliography}{10}
\providecommand{\url}[1]{\texttt{#1}}
\providecommand{\urlprefix}{URL }
\providecommand{\doi}[1]{https://doi.org/#1}

\bibitem{CAST2016}
{OSCT Borders X-ray Image Library, UK Home Office Centre for Applied Science
  and Technology (CAST)}. Publication Number: 146/16 (2016)

\bibitem{Abdallah2016}
Abdallah, A., Maarof, M.A., Zainal, A.: {Fraud detection system: A survey}.
  Journal of Network and Computer Applications  \textbf{68},  90--113 (jun
  2016). \doi{10.1016/J.JNCA.2016.04.007},
  \url{https://www.sciencedirect.com/science/article/pii/S1084804516300571}

\bibitem{Ahmed2016a}
Ahmed, M., Mahmood, A.N., Islam, M.R.: {A survey of anomaly detection
  techniques in financial domain}. Future Generation Computer Systems
  \textbf{55},  278--288 (feb 2016). \doi{10.1016/J.FUTURE.2015.01.001},
  \url{https://www.sciencedirect.com/science/article/pii/S0167739X15000023}

\bibitem{Ahmed2016}
Ahmed, M., {Naser Mahmood}, A., Hu, J.: {A survey of network anomaly detection
  techniques}. Journal of Network and Computer Applications  \textbf{60},
  19--31 (jan 2016). \doi{10.1016/J.JNCA.2015.11.016},
  \url{https://www.sciencedirect.com/science/article/pii/S1084804515002891}

\bibitem{Akcay2018}
Akcay, S., Kundegorski, M.E., Willcocks, C.G., Breckon, T.P.: Using deep
  convolutional neural network architectures for object classification and
  detection within x-ray baggage security imagery. IEEE Transactions on
  Information Forensics and Security  \textbf{13}(9),  2203--2215 (Sept 2018).
  \doi{10.1109/TIFS.2018.2812196}

\bibitem{an2015variational}
An, J., Cho, S.: Variational autoencoder based anomaly detection using
  reconstruction probability. Special Lecture on IE  \textbf{2},  1--18 (2015)

\bibitem{Arjovsky2017a}
Arjovsky, M., Bottou, L.: {Towards Principled Methods for Training Generative
  Adversarial Networks}. In: 2017 ICLR (April 2017),
  \url{http://arxiv.org/abs/1701.04862}

\bibitem{Arjovsky2017}
Arjovsky, M., Chintala, S., Bottou, L.: {W}asserstein generative adversarial
  networks. In: Proceedings of the 34th International Conference on Machine
  Learning. pp. 214--223. Sydney, Australia (06--11 Aug 2017),
  \url{http://proceedings.mlr.press/v70/arjovsky17a.html}

\bibitem{Chandola2009}
Chandola, V., Banerjee, A., Kumar, V.: {Anomaly detection}. ACM Computing
  Surveys  \textbf{41}(3),  1--58 (jul 2009). \doi{10.1145/1541880.1541882}

\bibitem{Chen2016}
Chen, X., Chen, X., Duan, Y., Houthooft, R., Schulman, J., Sutskever, I.,
  Abbeel, P.: {InfoGAN: Interpretable Representation Learning by Information
  Maximizing Generative Adversarial Nets}. In: Advances in Neural Information
  Processing Systems. pp. 2172--2180 (2016)

\bibitem{Creswell2016}
Creswell, A., Bharath, A.A.: Inverting the generator of a generative
  adversarial network (ii). arXiv preprint arXiv:1802.05701  (2018)

\bibitem{Creswell2017}
Creswell, A., White, T., Dumoulin, V., Arulkumaran, K., Sengupta, B., Bharath,
  A.A.: Generative adversarial networks: An overview. IEEE Signal Processing
  Magazine  \textbf{35}(1),  53--65 (2018)

\bibitem{Dimokranitou2017}
Dimokranitou, A.: {Adversarial Autoencoders for Anomalous Event Detection in
  Images}. Ph.D. thesis, Purdue University (2017)

\bibitem{Donahue2016}
Donahue, J., Kr{\"{a}}henb{\"{u}}hl, P., Darrell, T.: {Adversarial Feature
  Learning}. In: International Conference on Learning Representations (ICLR).
  Toulon, France (apr 2017), \url{http://arxiv.org/abs/1605.09782}

\bibitem{Dumoulin2016}
Dumoulin, V., Belghazi, I., Poole, B., Mastropietro, O., Lamb, A., Arjovsky,
  M., Courville, A.: Adversarially learned inference. In: ICLR (2017)

\bibitem{Goodfellow2014b}
Goodfellow, I., Pouget-Abadie, J., Mirza, M., Xu, B., Warde-Farley, D., Ozair,
  S., Courville, A., Bengio, Y.: Generative adversarial nets. In: Advances in
  neural information processing systems. pp. 2672--2680 (2014)

\bibitem{Gulrajani2017}
Gulrajani, I., Ahmed, F., Arjovsky, M., Dumoulin, V., Courville, A.C.: Improved
  training of wasserstein gans. In: Advances in Neural Information Processing
  Systems. pp. 5767--5777 (2017)

\bibitem{Hasan2016}
Hasan, M., Choi, J., Neumann, J., Roy-Chowdhury, A.K., Davis, L.S.: Learning
  temporal regularity in video sequences. In: Proceedings of the IEEE
  Conference on Computer Vision and Pattern Recognition. pp. 733--742 (2016)

\bibitem{Hodge2004}
Hodge, V., Austin, J.: {A Survey of Outlier Detection Methodologies}.
  Artificial Intelligence Review  \textbf{22}(2),  85--126 (oct 2004).
  \doi{10.1023/B:AIRE.0000045502.10941.a9},
  \url{http://link.springer.com/10.1023/B:AIRE.0000045502.10941.a9}

\bibitem{Ioffe2015}
Ioffe, S., Szegedy, C.: Batch normalization: Accelerating deep network training
  by reducing internal covariate shift. In: Proceedings of the 32nd
  International Conference on Machine Learning. pp. 448--456. Lille, France
  (07--09 Jul 2015), \url{http://proceedings.mlr.press/v37/ioffe15.html}

\bibitem{Isola2016}
Isola, P., Zhu, J., Zhou, T., Efros, A.A.: Image-to-image translation with
  conditional adversarial networks. In: 2017 IEEE Conference on Computer Vision
  and Pattern Recognition (CVPR). pp. 5967--5976 (July 2017).
  \doi{10.1109/CVPR.2017.632}

\bibitem{Kingma2014}
Kinga, D., Adam, J.B.: Adam: A method for stochastic optimization. In:
  International Conference on Learning Representations (ICLR). vol.~5 (2015)

\bibitem{Kiran2018}
Kiran, B.R., Thomas, D.M., Parakkal, R.: An overview of deep learning based
  methods for unsupervised and semi-supervised anomaly detection in videos.
  Journal of Imaging  \textbf{4}(2), ~36 (2018)

\bibitem{Krizhevsky2009}
Krizhevsky, A., Hinton, G.: Learning multiple layers of features from tiny
  images. Tech. rep., Citeseer (2009)

\bibitem{LeCun2010}
LeCun, Y., Cortes, C.: {MNIST} handwritten digit database  (2010),
  \url{http://yann.lecun.com/exdb/mnist/}

\bibitem{Lipton2017}
Lipton, Z.C., Tripathi, S.: Precise recovery of latent vectors from generative
  adversarial networks. In: ICLR Workshop (2017)

\bibitem{Makhzani2015}
Makhzani, A., Shlens, J., Jaitly, N., Goodfellow, I., Frey, B.: Adversarial
  autoencoders. In: ICLR (2016)

\bibitem{Markou2003a}
Markou, M., Singh, S.: {Novelty detection: a review—part 1: statistical
  approaches}. Signal Processing  \textbf{83}(12),  2481--2497 (dec 2003).
  \doi{10.1016/J.SIGPRO.2003.07.018},
  \url{https://www.sciencedirect.com/science/article/pii/S0165168403002020}

\bibitem{Markou2003}
Markou, M., Singh, S.: {Novelty detection: a review—part 2:: neural network
  based approaches}. Signal Processing  \textbf{83}(12),  2499--2521 (dec
  2003). \doi{10.1016/J.SIGPRO.2003.07.019},
  \url{https://www.sciencedirect.com/science/article/pii/S0165168403002032}

\bibitem{Medel2016}
Medel, J.R., Savakis, A.: {Anomaly Detection in Video Using Predictive
  Convolutional Long Short-Term Memory Networks}. CoRR  \textbf{abs/1612.0}
  (dec 2016)

\bibitem{Mirza2014}
Mirza, M., Osindero, S.: Conditional generative adversarial nets. arXiv
  preprint arXiv:1411.1784  (2014)

\bibitem{Paszke2017}
Paszke, A., Gross, S., Chintala, S., Chanan, G., Yang, E., DeVito, Z., Lin, Z.,
  Desmaison, A., Antiga, L., Lerer, A.: {Automatic differentiation in PyTorch}
  (2017)

\bibitem{Pimentel2014}
Pimentel, M.A., Clifton, D.A., Clifton, L., Tarassenko, L.: A review of novelty
  detection. Signal Processing  \textbf{99},  215--249 (2014)

\bibitem{Radford2015}
Radford, A., Metz, L., Chintala, S.: {Unsupervised Representation Learning with
  Deep Convolutional Generative Adversarial Networks}. In: ICLR (2016)

\bibitem{Ravanbakhsh2017a}
Ravanbakhsh, M., Sangineto, E., Nabi, M., Sebe, N.: {Training Adversarial
  Discriminators for Cross-channel Abnormal Event Detection in Crowds}. CoRR
  \textbf{abs/1706.0} (jun 2017), \url{http://arxiv.org/abs/1706.07680}

\bibitem{rogers2016automated}
Rogers, T.W., Jaccard, N., Morton, E.J., Griffin, L.D.: {Automated x-ray image
  analysis for cargo security: critical review and future promise}. Journal of
  X-ray science and technology (Preprint),  1--24 (2016)

\bibitem{Sabokrou2015}
Sabokrou, M., Fathy, M., Hoseini, M., Klette, R.: {Real-time anomaly detection
  and localization in crowded scenes}. 2015 IEEE Conference on Computer Vision
  and Pattern Recognition Workshops (CVPRW) pp. 56--62 (2015).
  \doi{10.1109/CVPRW.2015.7301284},
  \url{http://ieeexplore.ieee.org/document/7301284/}

\bibitem{Salimans2016}
Salimans, T., Goodfellow, I., Zaremba, W., Cheung, V., Radford, A., Chen, X.:
  Improved techniques for training gans. In: Advances in Neural Information
  Processing Systems. pp. 2234--2242 (2016)

\bibitem{Schlegl2017}
Schlegl, T., Seeb{\"{o}}ck, P., Waldstein, S.M., Schmidt-Erfurth, U., Langs,
  G.: {Unsupervised anomaly detection with generative adversarial networks to
  guide marker discovery}. Lecture Notes in Computer Science (including
  subseries Lecture Notes in Artificial Intelligence and Lecture Notes in
  Bioinformatics)  \textbf{10265 LNCS},  146--147 (2017).
  \doi{10.1007/978-3-319-59050-9_12}

\bibitem{Zenati2018a}
Zenati, H., Foo, C.S., Lecouat, B., Manek, G., Chandrasekhar, V.R.: Efficient
  gan-based anomaly detection. arXiv preprint arXiv:1802.06222  (2018)

\end{thebibliography}

\end{document}